\documentclass[sn-mathphys]{sn-jnl}

\jyear{2021}%

\theoremstyle{thmstyleone}%
\theoremstyle{thmstyletwo}%

\theoremstyle{thmstylethree}%

\begin{document}

\title[Open surgery multi-camera tool classification]{Open surgery tool classification and hand utilization using a multi-camera system}


\author*[1]{\fnm{Kristina} \sur{Basiev}}\email{kris.basiev@gmail.com}

\author[2]{\fnm{Adam} \sur{Goldbraikh}}\email{sgoadam@campus.technion.ac.il}

\author[3]{\fnm{Carla M} \sur{Pugh}}\email{cpugh@stanford.edu}

\author[1]{\fnm{Shlomi} \sur{Laufer}}\email{laufer@technion.ac.il}

\affil*[1]{\orgdiv{Faculty of Industrial Engineering and Management}, \orgname{Technion – Israel Institute of Technology}, \orgaddress{ \city{Haifa}, \postcode{3200003}, \country{Israel}}}

\affil[2]{\orgdiv{Applied Mathematics Department}, \orgname{Technion – Israel Institute of Technology}, \orgaddress{ \city{Haifa}, \postcode{3200003}, \country{Israel}}}

\affil[3]{\orgdiv{Department of Surgery}, \orgname{Stanford University School of Medicine}, \orgaddress{  \postcode{610101}, \state{Stanford}, \country{California}}}


\abstract{The abstract serves both as a general introduction to the topic and as a brief, non-technical summary of the main results and their implications. Authors are advised to check the author instructions for the journal they are submitting to for word limits and if structural elements like subheadings, citations, or equations are permitted.}
The abstract should briefly summarize the contents of the paper in
150--250 words. The abstract should be in a structured format (Purpose, Methods, Results, Conclusion) and should not exceed 250 words.

\abstract{\textbf{Purpose: } The goal of this work is to use multi-camera video to classify open surgery tools as well as identify which tool is held in each hand. Multi-camera systems help prevent occlusions in open surgery video data.  Furthermore, combining multiple views such as a Top-view camera covering the full operative field and a Close-up camera focusing on hand motion and anatomy, may provide a more comprehensive view of the surgical workflow. However, multi-camera data fusion poses a new challenge: a tool may be visible in one camera and not the other. Thus, we defined the global ground truth as the tools being used regardless their visibility.  Therefore, tools that are out of the image should be remembered for extensive periods of time while the system responds quickly to changes visible in the video.  \\ 
\textbf{Methods: }Participants (n=48) performed a simulated open bowel repair. A Top-view and a Close-up cameras were used.  YOLOv5 was used for tool and hand detection. A high frequency LSTM with a 1 second window at 30 frames per second (fps) and a low frequency LSTM with a 40 second window at 3 fps were used for spatial, temporal, and multi-camera integration. \\
\textbf{Results: }The accuracy and F1 of the six systems were: Top-view (0.88/0.88), Close-up (0.81,0.83), both cameras (0.9/0.9), high fps LSTM (0.92/0.93), low fps LSTM (0.9/0.91), and our final architecture the Multi-camera classifier(0.93/0.94).\\
\textbf{Conclusion: } By combining a system with a high fps and a low fps from the multiple camera array we improved the classification abilities of the global ground truth. \\}




\maketitle
\section{Introduction}\label{sec3}
 Recording activities in the operating rooms within and around the surgical site can have an enormous impact on the delivery of care. Recorded data can be used for surgical education, phase recognition \cite{zisimopoulos2018deepphase}, workflow analysis  \cite{primus2018frame}, error analysis \cite{xiao2007video}, skill assessment \cite{jin2018tool,partridge2014accessible},  and video summarization \cite{liu2020ultrasound}.
The process of estimating the objects' positions, sizes, and classifications is termed object detection. Tool and hand detection as well as identifying which tools are in each hand are fundamental tasks necessary for analyzing surgical video data.\cite{sznitman2011unified,richa2011visual}.

Typically, in minimal invasive surgery (MIS) one internal camera is always present. The camera serves as the surgeon's eyes and  is actively moved by the surgeon who dictates position and field of view. Thus, most of the time the camera is focused on the surgical working area. \\  
In contrast to MIS, open surgery requires the identification of more hands handling more tools. Using only one camera in open surgery presents several challenges. First, due to the dynamic nature of this working environment occlusions of surgical site are a constant issue. Second, in order to receive the full picture multiple view points are necessary. For example, a Close-up view may assist in identifying the different anatomical landmarks as well as help distinguish between similar tools. While a far away top view may offer a view of the tool tray as well as the assistant's hand \cite{shimizu2021hand}. 

In the computer vision field, the task of object detection is well studied    \cite{redmon2018yolov3,ren2015faster}. In the medical field, most studies focus on MIS \cite{richa2011visual,sznitman2011unified,al2018monitoring,jin2018tool}, with only few focusing on tool detection in open surgery\cite{shimizu2021hand,zhang2020using,goldbraikh2021videobased}.
Several works have used multi-angle photography during MIS, with the internal camera capturing the tools being used, and the other cameras identifying the area surrounding the surgery.  Using such systems enables detection of automatic activity \cite{schmidt2021multi},  and inoperative errors and distractions, as well as the evaluation of skill level \cite{jung2020first,ayas2021effect}. 
In another open surgery study, multiple cameras were placed in the surgical lamp \cite{kajita2020overhead,hachiuma2020deep}. Algorithms were developed to  automatically select the image with the best view of the surgical field.  This method generates a single video which is then analyzed. This approach helps deal with occlusions of the surgical site. However all cameras fundamentally provide the same viewing angle and same level of detail.\\

\begin{figure}
\begin{center}
\includegraphics[scale=0.5]{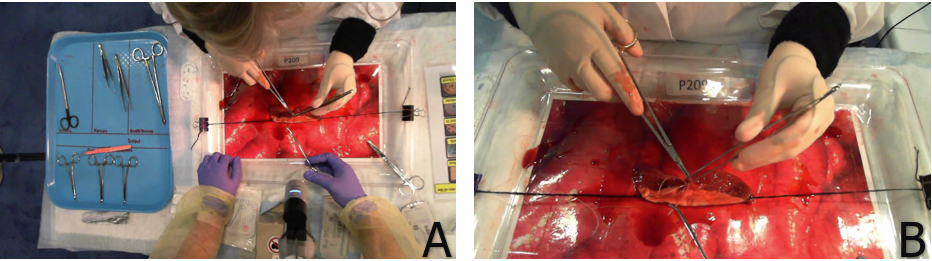}
\caption{Comparison between the images of Top-view camera (A) and Close-up camera (B).} \label{Img3}
\end{center}
\end{figure}

The present  study provides the analysis of videos captured during a simulated open bowel repair. We combined information from two cameras with two different viewing angles: a Top-view camera covering the full operative field and a Close-up camera focusing on hand motion and tools. These two views not only reduce the risk of site occlusion, but complement each other by providing differed fields of view and different image resolutions (Fig. \ref{Img3}).\\
However, using multiple views raises new challenges. The first challenge revolves around the basic definition of "ground truth". Traditionally, when tool classification is preformed we expect the algorithm to identify which tool is being used, only when the tool is in the video  image. However, when multiple cameras are involved the tool may be visible in one camera while  occluded or outside the field of view of the other camera. \\
Therefore, in this study we defined a new concept: the ‘global ground truth’. For example, if the surgeon's right hand is visible in at least one camera, the ‘global ground truth’ will be based on the camera in which it is visible. If it disappears for a short time from both cameras, we will assume the hand is still holding the tool it held when last visible. This provides consistent labeling for each camera and their combination. Therefore, we will add a memory component capable of remembering the tool held last. The only limitation of this approach is in case the tool is switched, and the switch is not visible. However, from our experience in most cases when the hand disappears from the image it reappears with the same tool in the same hand. Tool replacements account for a small fraction of the cases. \\
The need to remember tools that are out of the image, typically requires remembering extend time periods (20-30 seconds). On the other hand many events are considerably shorter. For example, the scissors are typically held for only few seconds. Furthermore, we expect that a temporal system will help filter classification noise (e.g. a single frame of tool B among a sequence of tool A). Therefore, our temporal system should address very different timescales. Finally, in addition to temporal issues, a system that combines two cameras should intelligently determine which tool is used when the cameras contradict each other.
When analyzing temporal information the amount of history evaluated as well as the sampling frequency play a vital role. A very long history might mask fast events while using a short history might loss track of tools out of the field of view. Sampling frequency, which in our case is measured by frames per second (fps), should typically correlate with the speed of the events analyzed. Furthermore, if a low fps is used a longer time window may be used while sustaining a reasonable input size. Several tool classification studies evaluated sequence length or specific fps to optimize results \cite{shimizu2021hand,liu2020anchor,kondo2021lapformer,peven2021activity}. In this study we chose a dual LSTM \cite{schmidhuber1997long} system. One system received 1 second of data at 30fps and the second 40 seconds at 3fps. The combination of both systems is able to balance between the fast events and the historical data.

To address these challenges, we developed a Multi-camera classifier (MCC) with the following two main components. The goal of the first component is to perform detection of all tool hand combinations. To this end we used the YOLOv5 \cite{yolov5}, which involved providing a tight bounding box for each hand and tool in the image of each camera. The next component integrates spacial-temporal information from both cameras, to provide the ‘global ground truth’ which classifies which tool is present in each hand. This was termed the classification phase.\\
To the best of our knowledge this study is the first study to perform tool classification in open surgery in a multi-view setup. The main contributions of this study are first a unified method for assessing the classification capabilities of each camera angle (termed ‘global ground truth’). Second, we developed a dual LSTM system for analyzing multiple timescales. Finally, we combined information from multi-view cameras to achieve optimal classification of a simulated open surgery procedure.

\section{Methods}\label{sec4}
\subsection{Simulator}\label{subsec2}

The data was collected using a simulator designed to represent a trauma patient in the operating room with an injured bowel. The data was collected in the exhibit hall of the 2019 American College of Surgeons (ACS) Clinical Congress annual meeting in San Francisco, CA. The simulator consisted of a porcine intestine arranged on a tray with surrounding artificial blood. There were ten stations in the area, each containing a tray of surgical instruments, sutures, absorbent pads, a step stool, a stopwatch, an overhead light and a tray containing a segment of porcine bowel that had a standardized injury. In each station, one camera was fixed on a frame to collect video data from above. A second camera was set up as a zoom-in to the surgeon’s hands and the intestine. Two full-thickness injuries in close proximity were located at the antimesenteric border. A trained researcher who is considered to be at the level of a medical student acted as a surgical assistant during the simulated procedure. Participants were required to decide how to repair the injury. They had 15 min to complete the surgical task. On average most participants finished the procedure in approximately 10  minutes. In addition, in order to assess the quality of the repair, a dedicated device was developed to examine if the repaired bowel leaked.  The study was approved by the Institutional Review Board at Stanford University. \\      

\subsection{Data collection}\label{subsec3}
Our total data-set includes videos from 2 angels and of approximately 200 participants (a total of about 400 videos).    Data from 28 participants (56 videos) were used for detection labeling. From that set of videos, 8300 frames were sampled for training, 2100 for validation, and 1100 reserved for the test set. Since the data was unbalanced and a few categories barely existed, the data was balanced as much as possible by using horizontal and vertical augmentations on all the images. The labeling was performed using Microsoft's Visual Object Tagging Tool. Each object was tagged with bounding boxes which indicate the location of the object as well as the type of the object inside each bounding box. The data included 20 classes: 2 surgeon hands and 2 assistant hands, each hand either empty or holding one of 4 tools.\\
To accomplish the classification task, the 'global ground truth' of 20 participants were labeled. This included start and end time of each event which was performed using BORIS (Behavioral Observation Research Interactive Software). Thus, for the classification task the full video was labeled and not just specific frames. The training and evaluation were performed on a NVIDIA Tesla V100 Volta GPU Accelerator 32GB Graphics Card.\\
\begin{table}[h]
\begin{center}
{\scalebox{.9}{
\begin{minipage}{\textwidth}
\caption{List of the classes defined in the data-set.}\label{tab1}%
\begin{tabular}{@{}llll@{}}
\toprule
Class ID & Surgeon's class &Class ID & Assistant's class \\
\midrule
\textbf{S}RE &  Empty right hand & \textbf{A}RE & Empty right hand \\
\textbf{S}RN &  Right hand with needle holder & \textbf{A}RN & Right hand with needle holder \\
\textbf{S}RF &  Right hand with forceps & \textbf{A}RF &Right hand with forceps \\
\textbf{S}RS &  Right hand with scissors & \textbf{A}RS & Right hand with scissors \\
\textbf{S}RM &  Right hand with mosquito forceps & \textbf{A}RM & Right hand with mosquito forceps \\
\textbf{S}LE &  Empty left hand & \textbf{A}LE & Empty left hand \\
\textbf{S}LN &  Left hand with needle holder & \textbf{A}LN & Left hand with needle holder \\
\textbf{S}LF &  Left hand with forceps & \textbf{A}LF & Left hand with forceps \\
\textbf{S}LS &  Left hand with scissors & \textbf{A}LS & Left hand with scissors \\
\textbf{S}LM &  Left hand with mosquito forceps & \textbf{A}LM & Left hand with mosquito forceps \\
\botrule
\end{tabular}
\end{minipage}
}
}
\end{center}
\end{table}
\subsection{Detection}\label{subsec4}
First, the detection model was trained to detect hands and tools in each of the two angles. The detection identified one of four hands: surgeon or assistant and left or right. In addition it identified one of five categories: empty, needle holder, forceps and scissors. Thus the model was trained to identify 20 classes. For the detection task, we used the YOLOv5 network (YOLOv5x). The same network was trained for both angles.\\
YOLOv5 is a real-time object detector. While real-time was not the objective of this study, providing performance feedback is one of the long term goals of this project, and fast analysis is of great value when feedback is wanted. The algorithm outputs a detection bounding box with the object's class, objectness and coordinates.\\
YOLO uses Non-Maximum Suppression (NMS) to deal with multiples of the same object. The network was trained by using a SGD optimizer with a learning rate of 0.01. Leaky ReLU activation function was used in the middle and on the hidden layers and the sigmoid activation function was used in the final detection layer. The model trained for 150 epochs. The model chosen exhibited the best mAP performance, based on the validation data.\\
Finally, for the classification step, 40 new videos are used. These videos were analyzed using the detection system. That is, the output of the detection serves at the input of the classification.\\

\subsection{Naive Approach}\label{subsec5}
The goal of the classification task is to identify what is being held in each of the four hands involved in the simulation (the surgeon's and the assistant's). Thus the system should always consist of four outputs, one for each hand. Each output receives one of five classes: one of four tools or empty. For the 'global ground truth' classification, our aim is to classify in every frame all four hands even if some hands or tools are visible only at one angle or not visible at all.\\
As a baseline we first present a naive approach for predicting the 'global ground truth'. YOLO's output contains muliple objects. For each object it provides its class, position of the bounding box and the probability. These data are provided independently for each camera and depend only on the current frame. A number of guidelines have been set up to classify whether the hand is empty or holds one of the tools: whenever there are contradicting classes detected for any hand, we choose the one that has the highest probability from both angels. In the case where the hand is detected only by one camera, we choose the class with the highest probability from the camera that detects the hand. When the hand isn't detected by both cameras, the previous detection is used, assuming that whatever was in the surgeon's hand was likely to remain there until we see otherwise.\\

\subsection{Multi-Camera Classifier Approach}\label{subsec6}

\begin{figure}
\centering
\includegraphics[scale=0.35]{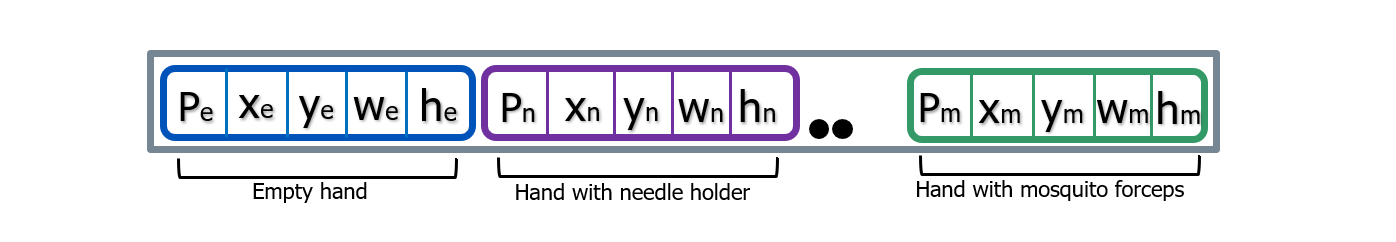}
\caption{Schematic illustration of the input vector of each hand.} \label{vector25}
\end{figure}

Our architecture was developed to achieve the most accurate and smooth classification by combining detection data from both angles as well as temporal information. The detection algorithm goal was to identify the type and position of all the tools and hands in both videos. In addition, the system should take into account the that some events are very fast while some are very slow.   
As mentioned, for each detection bounding box, YOLO's output included: the object class, the class probability, and the four coordinates of the bounding box . When multiple objects were detected for the same hand, the object with the highest probability was selected. 
The input for the LSTM was a vector with 25 elements for each camera and each hand (Fig. \ref{vector25}).\\ 
For each hand, the vector included a concatenation of five detection vector, one for each of the five classes. In each detection vector, $p_{i}$ is the tools probability. This is followed by the 4 elements for the position coordinates of the bounding box, $x_{i}$  and $y_{i}$ represent the center and $w_i$ and $h_i$  represent width and height respectfully. Zeros complete the class vector when there is no optional class detection. \\
There are a number of constraints regarding the duration of the temporal information. Long sections of information from the past could override tools that have been in use for a short time. By using only a sequence of close samples, the history required long periods of time in which the hands are out of the frame, will be missed. Thus we decided to use two LSTMs.  The input of the first LSTM is 1 second of data at 30 fps, for the short and quick events. The input of the second LSTM was 40 seconds of data at 3 fps, for historical events. \\
Therefore, we compared each LSTM separately as well as their combination. For combining the LSTMs, the last hidden layer (length 32) of all LSTMs was concatenated and was passed through two fully connected layers with Relu activation function. Finaly, to predict the classification probability function softmax activation function was applied to the output layer. The model was trained for 2000 epochs with a learning rate of 1e-4, dropout probability of 0.3 and batch size 16.

\subsection{Evaluation Metrics}\label{subsec8}
In order to evaluate and compare the model performance, we chose to use some familiar metrics:  accuracy (ACC), precision (PR), recall (RE) and F1 score (F1). Accuracy quantities the total correct classifications of the 'global ground truth'.
For each class predictions and 'global ground truth' PR, RE, and F1 were computed.\\
Since our data is unbalanced, when summarizing the results on all classes several variants of the F1 score were calculated:\\
F1 weighted - calculates the weighted average of F1. Where the weight of each of the 20 classes is the number of instances of that class.\\
F1 macro -  calculates the unweighted average of F1. That is, each class has the same weight regardless its prevalence. This score emphasis classes with small representations.\\
F1 macro for more than 100 or 200 - Similar to F1 macro, but first ignores class with less than 100 of 200 instances, receptively. This approach ignores extremely small classes which most likely don't have enough data for the system to learn.

\begin{figure}
\includegraphics[width=\textwidth]{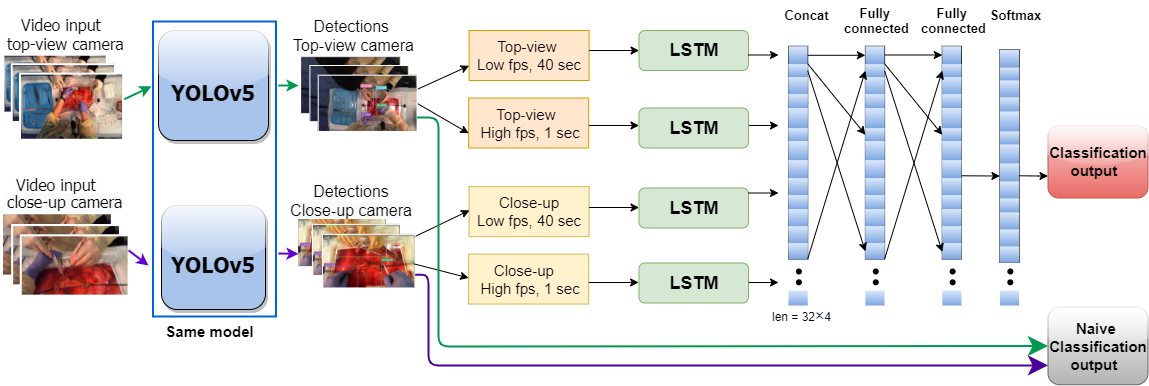}
\caption{Schematic illustration of the Multi-camera classifier (MCC).} \label{DETECTION_4}
\end{figure}

\section{Results}\label{sec5}
The performance detection component is represented at Table \ref{tab2}. The results are based on the test set (1100 images). The system yielded an expectation mAP of 0.764. To test the robustness of the detection algorithm, we performed the experiments via the simulation-out setting, in which the images for training and the images for testing are from different videos.
As shown in the table, there are classes that do not exist in the test set since their occurrence is rare. \\
\begin{table}[h]
\begin{center}
{\scalebox{.85}{
\begin{minipage}{160pt}
\caption{Average Precision (AP) detection test results.}\label{tab2}%
\begin{tabular}{@{}lllllll@{}}
\toprule
Class ID & Occurrence & $AP_{50}$ \\\midrule
SRE &  203 & 0.922  \\
SRN &  689 & 0.975  \\
SRF &  136 & 0.866 \\
SRS &  41 & 0.766\\
SLE &  347 & 0.948  \\
SLF & 755 & 0.981  \\
ARE &  259 &  0.841  \\
ARS &  206 & 0.943 \\
ARM &  155 & 0.826 \\
ALE &  380 & 0.898\\
ALM &  160 & 0.889 \\
\hline
Average &  3331 & 0.764 \\
\end{tabular}
\end{minipage}
}
}
\footnotetext{Note: Results for classes with occurrence in the test set are shown in the table. In the following classes, no results were obtained due to no shows: SRM, SLN, SLS, SLM, ARN, ARF, ALN, ALF and ALS.}
\end{center}
\end{table}
For the classification task, we analyzed performance via the simulation-out setting, in which we train our model on fifteen surgery videos and test it on five videos. The values were calculated for the 4-fold cross-validation set, and then the mean of the set was calculated. \\
In Table \ref{tab3} the scores for each camera are compared independently to the naive approach as well as with our final architecture MCC. MCC outperform the naive solution. In addition, high-frequency LSTM (30fps), low-frequency LSTM (3fps) and the MCC compered. In most classes, the low-frequency LSTM outperformed the high-frequency LSTM. MCC provided the best results\\
\begin{table}[h]{\scalebox{.85}{
\begin{minipage}{395pt}
\caption{The video classification results for each of the evaluated models.}\label{tab3}
\begin{tabular*}{\textwidth}{@{\extracolsep{\fill}}lcccccccccccccccc@{\extracolsep{\fill}}}
\toprule%
& \multicolumn{1}{@{}c@{}}{ } & \multicolumn{3}{@{}c@{}}{Naive approach} & \multicolumn{3}{@{}c@{}}{High fps}& \multicolumn{3}{@{}c@{}}{Low fps}& \multicolumn{3}{@{}c@{}}{MCC} \\
\cmidrule{3-5}\cmidrule{6-8}\cmidrule{9-11}\cmidrule{12-14}%

  \multicolumn{1}{@{}c@{}}{\centering Class}
& \multicolumn{1}{p{1cm}}{\centering Num \\ Frames}
& \multicolumn{1}{@{}c@{}}{\centering PR}
& \multicolumn{1}{@{}c@{}}{\centering  RE}
& \multicolumn{1}{@{}c@{}}{\centering F1}
& \multicolumn{1}{@{}c@{}}{\centering PR}
& \multicolumn{1}{@{}c@{}}{\centering  RE}
& \multicolumn{1}{@{}c@{}}{\centering F1}
& \multicolumn{1}{@{}c@{}}{\centering PR}
& \multicolumn{1}{@{}c@{}}{\centering  RE}
& \multicolumn{1}{@{}c@{}}{\centering F1}
& \multicolumn{1}{@{}c@{}}{\centering PR}
& \multicolumn{1}{@{}c@{}}{\centering  RE}
& \multicolumn{1}{@{}c@{}}{\centering F1}\\

\midrule
SRE &  4346&  0.82&  0.94 & 0.88 &  0.87 & 0.93&  0.9 & 0.92&  0.89 & 0.9&  0.92 & 0.92&  0.92  \\
SRN &  12921 & 0.94&  0.89 & 0.91&  0.96 & 0.82&  0.88 & 0.95&  0.91 & 0.93&  0.95 & 0.92&  0.94  \\
SRF &  2113 & 0.9&  0.8 & 0.85  &  0.83 & 0.89&  0.85 & 0.84&  0.9 & 0.86&  0.85 & 0.91&  0.88  \\
SRS &  850 & 0.42&  0.51 & 0.43&0.33 & 0.69&  0.43 & 0.48&  0.63 & 0.5&  0.52 & 0.63&  0.52  \\
SRM &  115 & 0.16 & 0.05 & 0.08  &   0.3 & 0.05&  0.03 & 0.15 & 0.09 & 0.09&  0.13 & 0.04&  0.05  \\
SLE &  6315& 0.86&  0.95 & 0.89 &  0.89 & 0.92&  0.9 & 0.92&  0.95 & 0.93&  0.93 & 0.95&  0.94  \\
SLN &  294 & 0.21&  0.28 & 0.19&  0.04 & 0.2&  0.06 & 0.26&  0.21 & 0.19&  0.27 & 0.22&  0.23  \\
SLF &  13699 & 0.96&  0.93 & 0.95&   0.96 & 0.91&  0.94 & 0.96&  0.96 & 0.96&  0.97 & 0.97&  0.97  \\
SLS &  2   0 &  0 & 0 &  0 & 0&  0 & 0&  0 & 0&  0 & 0&  0  \\
SLM &  35& 0.14&  0.01 & 0.02 &   0 & 0&  0 & 0&  0 & 0&  0 & 0&  0  \\
ARE &  16045& 0.96&  0.95 & 0.96 &   0.98 & 0.97&  0.97 & 0.98&  0.96 & 0.97&  0.98 & 0.97&  0.97  \\
ARN &  106   & 0.12&  0.05 & 0.06&  0 & 0&  0 & 0&  0.09 & 0.04&  0 & 0.05&  0.04  \\
ARF &  2091  & 0.75&  0.6 & 0.66 & 0.83 & 0.80&  0.81 & 0.8&  0.86 & 0.83&  0.82 & 0.87&  0.83  \\
ARS &  384   & 0.2&  0.57 & 0.27&  0.1 & 0.28&  0.14 & 0.22&  0.41 & 0.27&  0.2 & 0.25&  0.2  \\
ARM &  1719 & 0.69&  0.52 & 0.49 & 0.77 & 0.61&  0.66 & 0.76&  0.73 & 0.74&  0.68 & 0.7&  0.68  \\
ALE &  16952& 0.95&  0.98 & 0.96 &  0.98 & 0.97&  0.98 & 0.97&  0.96 & 0.97&  0.97 & 0.98&  0.98  \\
ALN &  39& 0.07&  0.04 & 0.05 &  0 & 0&  0 & 0&  0 & 0&  0 & 0&  0  \\
ALF &  702 & 0.55&  0.34 & 0.39 &   0.48 & 0.44&  0.46 & 0.35&  0.4 & 0.37&  0.65 & 0.49&  0.45  \\
ALS &  113 & 0.26&  0.5 & 0.32&   0.26 & 0.65&  0.37 & 0.23&  0.57 & 0.32&  0.32 & 0.52&  0.34  \\
ALM &  2539  & 0.71&  0.63 & 0.6 &   0.73 & 0.7&  0.7 & 0.71&  0.77 & 0.7&  0.71 & 0.68&  0.7  \\
\botrule
\end{tabular*}
\end{minipage}
}
}
\end{table}
Table \ref{tab5} presents the average Accuracy and F1 of all models. The Multi-camera classifier outperforms the naive approach by 3\% in terms of Accuracy, 4\%
in terms of F1 weighted, 6\% for F1 score macro more than 200 images. Moreover, the Multi-camera classifier outperforms both single LSTM systems by 1-3\% with respect to Accuracy and 1-4\% for F1.
\begin{table}[h]{\scalebox{.8}{
\begin{minipage}{\textwidth}
\caption{Video classification Results.}\label{tab5}%
\begin{tabular}{@{}lcccccc@{}}
\toprule

& \multicolumn{1}{p{1.5cm}}{\centering Top-view \\  Camera}
& \multicolumn{1}{p{1.5cm}}{\centering Close-up \\  Camera}
& \multicolumn{1}{p{1.5cm}}{\centering Naive \\ approach}
& \multicolumn{1}{p{1.5cm}}{\centering Low fps}
& \multicolumn{1}{p{1.5cm}}{\centering High fps}
& \multicolumn{1}{p{2cm}}{\centering MCC}\\

\midrule
Accuracy &  0.88 & 0.81 &  0.9 & 0.9 &   0.92 & 0.93 \\
F1 Score Weighted &  0.88 & 0.83 &  0.9 & 0.91 & 0.93  & 0.94 \\
F1 Score Macro &  49 & 0.41 &  0.5 & 0.5 & 0.53  & 0.53 \\
F1 Score Macro more\\ than 100 &  0.63 & 0.55 &  0.64 & 0.66& 0.68& 0.7 \\
F1 Score Macro more\\ than 200 &  0.71 & 0.64 &  0.73 &  0.75&  0.78 & 0.79 \\
\botrule
\end{tabular}

\end{minipage}
}
}
\end{table}
\section{Discussion}\label{sec6}
In this paper, we studied different methods for using multi-view video data to classify the tools used during an open surgery simulation. Since each camera may have a partial view of the procedure, we defined a 'global ground truth'. The requirement of one unified output is obvious. One cannot have a system that declares different tools used based on the camera evaluated. By defining 'global ground truth' at the data labeling phase we help the system focus on this required result at the learning phase, eliminating the need for any heuristic decisions.\\ 
Examining the data revealed that while the LSTM systems were developed to analyze long and short memory (as declared in their name), the LSTM are sensitive to sample size as well as to sampling frequency. Thus, when multiple timescales exist one LSTM might not be ideal for capturing all the events. Therefore, the data was down-sampled to 3 fps. After down-sampling a 40 second sequence only include 120 samples, which is a reasonable input size. In addition a 1 second sequence with a sampling frequency of 30 fps was also used by an additional LSTM.\\
One challenge was that the fact the data were unbalanced. Thus, several classes were too small to classify. Efficient ways for expanding sample size should be developed.\\
In this study, by labeling only a fraction of the data, we were able to develop a tool that identified which tools were used throughout the procedure. This may facilitate a more detailed analysis surgical skill and technique or assist in action recognition \cite{peven2021activity}. In addition, the methods developed may be used for analyzing more complex data from the operating room.  

\section*{Declarations}
The work was supported by the American College of Surgeons (Stanford Sponsored Project \# 162769) and National Institutes of Health (1R01DK12344501A1). 
The authors would like to thank  Hossein Mohamadipanah,  LaDonna Kearse, Anna Witt, Brett Wise, Su Yang, Cassidi Goll, Anton Kanarski and Michal Tal. For assisting in data collection and annotation. 



\bibliography{ref.bib}

\end{document}